\documentclass[conference]{IEEEtran}
\usepackage{times}

\usepackage[numbers]{natbib}
\usepackage{booktabs}
\usepackage{array}
\usepackage{amsmath}
\usepackage{graphicx}
\usepackage{multirow}
\usepackage{caption}
\usepackage{placeins}
\usepackage{dblfloatfix}
\PassOptionsToPackage{hyphens}{url}
\usepackage[bookmarks=true]{hyperref}

\newcolumntype{L}[1]{>{\raggedright\arraybackslash}p{#1}}
\hypersetup{
   pdfauthor={Ke Rui, Yushen Zuo, Jiawei Wang, Haoran Jia, Jinming Ma, Weitao Zhou, Minglei Li},
   pdftitle={Diagnosing Semantic Handoff Failures in Agent-Orchestrated Vision-Language-Action Skill Composition},
   pdfsubject={Semantic Robotics},
   pdfkeywords={Semantic Robotics; Vision-Language-Action; Skill Composition; Verification}
}

\begin{document}

\title{Diagnosing Semantic Handoff Failures in Agent-Orchestrated Vision-Language-Action Skill Composition}

\author{
Ke Rui\textsuperscript{*}, Yushen Zuo\textsuperscript{*}, Jiawei Wang\textsuperscript{*\dag}, Haoran Jia, Jinming Ma, Weitao Zhou, Minglei Li\textsuperscript{\dag}\\[3pt]
SimpleAI\\[3pt]
\textsuperscript{*}Equal contribution\quad\textsuperscript{\dag}Corresponding authors: \texttt{\{wangjiawei, liminglei\}@simpleai.tech}
}

\makeatletter
\let\@oldmaketitle\@maketitle
\renewcommand{\@maketitle}{
\@oldmaketitle
\centering
\includegraphics[width=\textwidth]{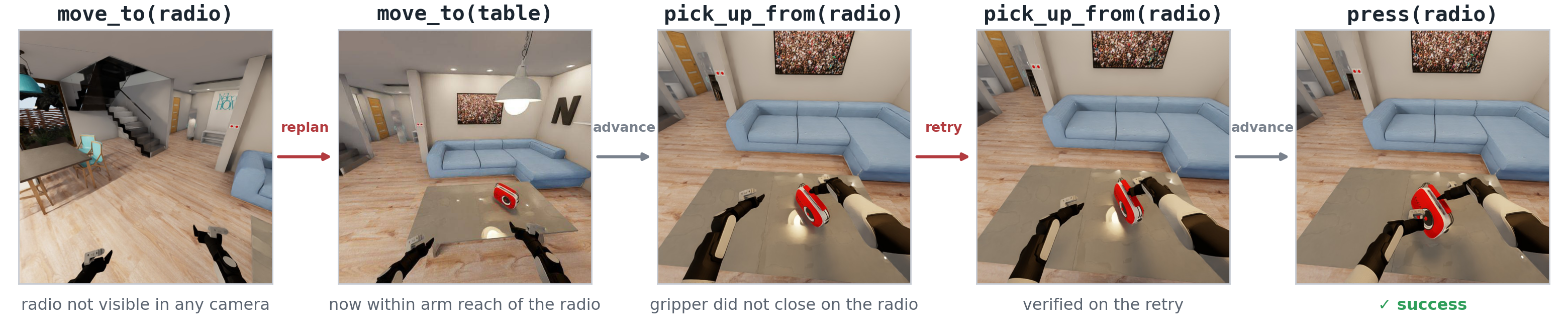}
\captionof{figure}{\textbf{Turning on the radio through skill composition.} The agent issues a sequence of typed skill calls such as \texttt{move\_to}, \texttt{pick\_up\_from}, and \texttt{press}, and a multi-view VLM verifier checks each handoff before the next skill runs. When a check fails the agent recovers within the same loop---re-planning to a new sub-goal after a navigation that does not reach the radio, and retrying after a grasp that does not close---advancing only once the post-condition holds, until the radio turns on. This plan-act-verify-replan loop is our agent harness; Fig.~\ref{fig:system-overview} gives the full architecture.}
\label{fig:teaser}
\addtocounter{figure}{-1}
}
\makeatother

\maketitle
\addtocounter{figure}{1}

\begin{abstract}
Long-horizon household tasks require robots to compose many language-conditioned skills, but the boundary between two skills is rarely explicit. A skill may satisfy its own postcondition while leaving the robot, objects, or camera views in a state from which the next skill cannot start. We study this \emph{semantic handoff} problem in BEHAVIOR-1K through an agent-orchestrated vision-language-action execution harness. The harness calls \(\pi_{0.5}\)-based skill checkpoints trained from cleaned BEHAVIOR-1K demonstrations, assigns each skill typed arguments and a step budget, and uses multi-view VLM verification to decide whether execution should advance, retry, or replan. To separate isolated skill competence from long-horizon composition, we compare the same checkpoints from clean skill-boundary snapshots and from chained terminal states produced by previous skills. Selected navigation, grasping, placement, and door-opening skills reach 77--100\% success from snapshots under human-reviewed verification, yet composed rollouts still stall from chained states. The resulting traces attribute failures to next-skill readiness, target grounding, and control execution, turning near-zero task success into actionable diagnostics for what VLA skill libraries must learn next: robustness to the messy chained-state distribution that clean demonstrations underrepresent.
\end{abstract}

\IEEEpeerreviewmaketitle

\section{Introduction}

Long-horizon household tasks require robots to maintain semantic state across many visually grounded actions. Making microwave popcorn, for example, means navigating to the kitchen, opening the microwave, inserting the bag, closing the door, and activating the appliance. No single language-conditioned action solves such a task; it requires composing many skills while preserving the conditions that make the next skill executable.

A natural architecture is to let an agent orchestrate vision-language-action (VLA) skills. In this view, VLA policies act as general-purpose, language-conditioned visuomotor tools, while the agent layer handles task decomposition, state tracking, verification, recovery, and evidence collection. Compared with hand-engineered task and motion planners or symbolic tool APIs\citep{liang2023code,fu2026cap}, this boundary promises broader coverage: new objects and appliances need not each come with a new symbolic controller. The challenge is that learned skills expose a less explicit interface. They may complete the local behavior requested by the agent without producing a state that is usable by the next skill.

We call this the \emph{semantic handoff} problem (Sec.~\ref{sec:handoff}). A skill can satisfy its own postcondition yet leave the robot, objects, or camera views in a state from which the following skill cannot start. For example, a navigation skill before a grasp should not be considered complete merely because the target object is visible somewhere in the scene; the object should be close enough and positioned plausibly for grasping. Similarly, opening a door may satisfy an open-door postcondition while moving the next target out of view, creating a downstream grounding failure. Thus, long-horizon execution requires judging not only whether the current skill appears complete, but whether the resulting state is ready for the next skill.

We study this problem in BEHAVIOR-1K~\citep{li2023behavior} using a compact \(\pi_{0.5}\)-based VLA skill library trained from cleaned skill demonstrations and an agent execution harness that performs bounded plan-act-verify-replan control. The harness represents each skill call with typed arguments, a step budget, and a handoff-aware postcondition; executes the corresponding VLA checkpoint; verifies progress from head and wrist observations with a multi-view VLM verifier; and records replayable traces of skill attempts, verifier decisions, replans, and final task predicates. Our goal is not to claim state-of-the-art BEHAVIOR-1K task success. Instead, we ask what such an agent harness can reveal when long-horizon rollouts fail.

This diagnostic framing separates isolated skill competence from compositional robustness. We evaluate the same skill checkpoints both from clean skill-boundary snapshots and from chained terminal states produced by previous skills. Several skills succeed frequently from curated snapshots under human-reviewed verification, yet composed rollouts still stall from the messy states produced during execution. The resulting traces turn near-zero task success into actionable evidence: failures can be attributed to next-skill readiness, target grounding, or control execution from chained states, pointing to the data and verification capabilities that VLA skill libraries must improve.

This paper makes three contributions:
\begin{itemize}
    \item We formulate long-horizon agent-orchestrated VLA execution as semantic skill composition, where each skill call has typed arguments, a budget, an expected postcondition, and an implicit next-skill readiness requirement; we identify semantic handoff failures as cases where a skill appears complete but does not leave the state ready for the next skill.
    \item We instantiate this formulation as a BEHAVIOR-1K semantic execution harness: \(\pi_{0.5}\)-based skill checkpoints trained from cleaned skill segments are invoked through a bounded plan-act-verify-replan loop with multi-view VLM verification, handoff-aware postconditions, recovery, and replayable trace artifacts.
    \item We provide trace-backed diagnostics showing that snapshot-state skill competence does not by itself carry over to chained initial conditions. On a representative BEHAVIOR-1K rollout round, the harness attributes failed skill attempts to next-skill readiness, target grounding, and control execution from chained states; a controlled arm-reach verifier ablation surfaces additional readiness failures and triggers more re-navigation attempts.
\end{itemize}

\section{Semantic Handoff Problem}
\label{sec:handoff}

Long-horizon execution chains language-conditioned skills, but the boundary between two skills is under-specified. Each skill is trained and benchmarked to reach its own postcondition \(\phi_t\) from a curated skill-boundary state; in composition, however, the next skill \(s_{t+1}\) must start from whatever state the previous skill \(s_t\) actually leaves behind. A skill can therefore satisfy \(\phi_t\) and still strand the robot, objects, or camera views in a state from which \(s_{t+1}\) cannot execute---a \emph{semantic handoff} failure. The agent must judge not merely whether the current state is a plausible completion of \(s_t\), but whether it is a valid \emph{starting} state for \(s_{t+1}\).

A standard verifier checks only \(\phi_t\), which is insufficient at a boundary. For instance, a \texttt{move\_to(can)} step followed by \texttt{pick\_up\_from(can)} should not advance merely because the can is visible at a distance; it should advance only when the can is also plausibly reachable by the gripper. The same checkpoint is also exercised on different inputs in the two settings: a single-skill benchmark evaluates \texttt{move\_to} from a restored demonstration snapshot, whereas an end-to-end rollout invokes it after the robot has navigated, opened doors, and shifted viewpoints, so the next skill no longer starts from the curated skill-boundary distribution. This competence drop at skill boundaries is a distribution shift in the imitation-learning sense~\citep{ross2011dagger}, but distinct from classic within-rollout compounding error: each skill's legitimate terminal state is already out of distribution for the \emph{next} skill---a shift between two composed policies rather than within a single one.

In the current system we approximate this handoff requirement directly in the skill postconditions rather than through a separate predicate: the natural-language postcondition authored for each \texttt{move\_to} step requires the target to be shown close-up and at arm-reach distance, not merely visible. This is a deliberately limited instantiation; a principled, next-skill-indexed readiness predicate \(\rho(s_t,s_{t+1})\) verified jointly with the postcondition, \(\phi_t \wedge \rho\), is a design direction we develop in Sec.~\ref{sec:future}.

\section{Semantic Execution Harness}
\label{sec:system}

\begin{figure*}[t]
\centering
\includegraphics[width=\textwidth]{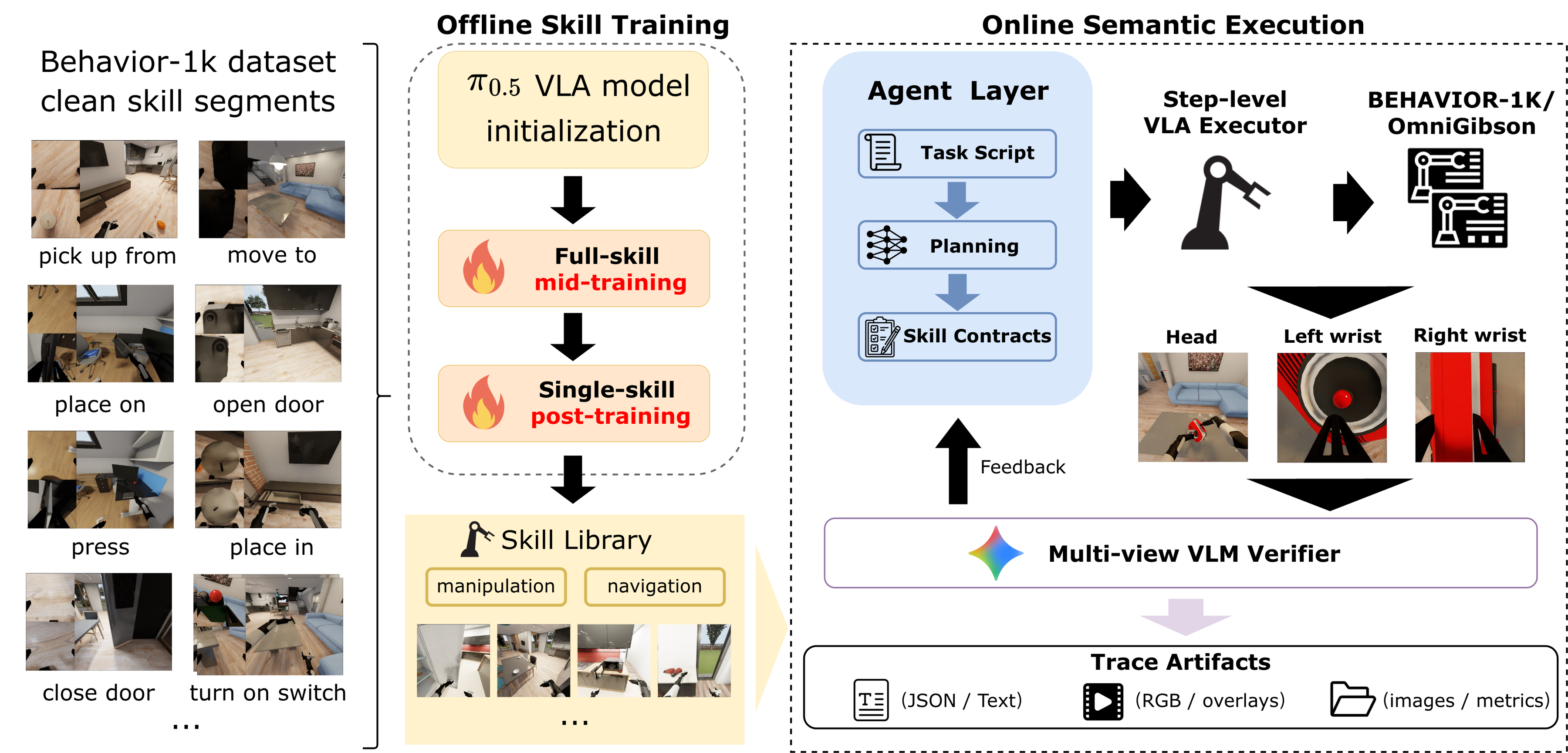}
\caption{\textbf{The semantic execution harness.} \emph{Offline} (left), cleaned BEHAVIOR-1K demonstrations initialize a \(\pi_{0.5}\) VLA backbone that full-skill mid-training and single-skill post-training specialize into a skill library spanning manipulation and navigation. \emph{Online} (right), the agent layer maintains a task script, planning, and typed skill contracts, and dispatches step-level VLA skills in BEHAVIOR-1K/OmniGibson; a multi-view VLM verifier judges progress from head and wrist views and feeds its verdict back to the agent to advance, retry, or replan. Every rollout writes replayable trace artifacts (logs, videos, and evidence).}
\label{fig:system-overview}
\end{figure*}

Our harness separates the agent layer from the embodiment layer. The agent layer selects skill calls, maintains task state, invokes verification, and triggers recovery. The embodiment layer owns simulation, VLA inference, and low-level action execution. Figure~\ref{fig:system-overview} summarizes the system.

\subsection{Semantic Skill Contracts}

We represent a skill call as
\[
s_t = (\text{name}, \text{args}, \text{prompt}, B_s, K_s, \phi_t),
\]
where \(\text{name}\) selects a skill type, \(\text{args}\) bind task objects, \(\text{prompt}\) is the language instruction passed to the VLA policy, \(B_s\) is the maximum skill budget, \(K_s\) is the verifier interval, and \(\phi_t\) is the expected postcondition. The current skill library includes \texttt{move\_to}, \texttt{pick\_up\_from}, \texttt{place\_on}, \texttt{place\_on\_next\_to}, \texttt{place\_in}, \texttt{press}, \texttt{turn\_on\_switch}, \texttt{open\_door}, and \texttt{close\_door}. We use these names throughout; trace artifacts label \texttt{move\_to} and \texttt{pick\_up\_from} as \texttt{navigate\_to} and \texttt{grasp}. \texttt{press} and \texttt{turn\_on\_switch} are distinct skills that share one checkpoint group: a button/switch press versus a toggle-on action.

The VLA executor is trained as part of the system rather than assumed as an external oracle: the skills are \(\pi_{0.5}\) vision-language-action policies~\citep{black2024pi0}, a PaliGemma vision-language backbone with a flow-matching action expert over three RGB views from the head and two wrists, plus proprioceptive state. We clean BEHAVIOR-1K teleoperation demonstrations~\citep{li2023behavior} into skill segments and train in two stages: a pooled all-skill mid-training, then skill-group specialization, with each skill type routed to its group's checkpoint at deployment (Table~\ref{tab:skill-checkpoints}). We use skill-specific budgets \(B_s\) because VLA policies lack a reliable stop signal and BEHAVIOR-1K skill durations vary widely (Table~\ref{tab:skill-duration}).

\subsection{Step-Level Plan-Act-Verify-Replan}

For each skill, the harness executes a VLA action chunk for \(K_s\) environment steps, collects head and wrist observations, and queries a VLM verifier over the head, left-wrist, and right-wrist views together with the current skill's postcondition. The skill advances as soon as a verification returns a positive verdict above a confidence threshold of \(0.6\); an \texttt{uncertain} verdict never terminates a skill, a fail-safe against flaky calls. Otherwise execution continues until the skill budget is exhausted or the agent replans. Because the agent, not the VLA, owns termination, every rollout produces a step-resolved verifier trajectory rather than a single end-of-skill verdict.

\section{Diagnostic Evaluation}

\subsection{Setup}

We evaluate in BEHAVIOR-1K with two complementary protocols: an isolated \emph{single-skill} benchmark and a composed \emph{end-to-end} rollout. The single-skill benchmark restores the simulator snapshot at a cleaned demonstration segment's start frame from the validation split and runs the corresponding skill-group checkpoint in isolation, scoring success with the multi-view VLM verifier whose decisions are confirmed by human review. These snapshots are drawn from the same task instances we evaluate end-to-end, so the isolated-vs-composed comparison holds the scene and instance fixed and varies only the skill's initial-state distribution---a clean snapshot versus the chained terminal state of the previous skill. The end-to-end protocol composes the same checkpoints through the agent loop under live VLA execution, each skill starting from the terminal state of the previous one; we run official BEHAVIOR-1K public test instances per task, taken from the 2025 challenge \texttt{test\_instances.csv}, one instance per round across three rounds. Execution is step-level plan-act-verify-replan: the agent verifies every \(K=200\) simulator steps with \texttt{gemini-2.5-flash} multi-view verification and advances a skill only when its postcondition holds, including an arm-reach handoff criterion on navigation, terminating on a per-skill step budget and a global episode budget of twice the human-demonstration length. Every rollout writes a structured trace of skill dispatches, verifier decisions, replans, final task-predicate checks, and media, which we parse into per-skill attempts; failed attempts are classified from structured verifier reasons as diagnostic labels, not human ground truth. The progress score aggregates all three rounds (\(n{=}3\)); the per-attempt failure attribution and verifier-call diagnostics are reported on one representative round.
\subsection{Several Skills Are Competent From Curated Snapshot States}

Table~\ref{tab:single-skill} summarizes the isolated skill benchmark. Several skills achieve strong judge success from curated skill-boundary states: \texttt{place\_on} reaches 6/6, \texttt{open\_door} reaches 7/7, \texttt{pick\_up\_from} reaches 28/29, \texttt{place\_in} reaches 11/13, and \texttt{move\_to} reaches 27/35. Notably, the placement skills---which dominate the end-to-end bottlenecks below---are themselves competent from snapshot states once trained to convergence, as Table~\ref{tab:skill-checkpoints} shows, so their chained failures cannot be attributed to weak placement policies in isolation. The skill library is therefore not uniformly weak in isolation.

\begin{table}[t]
\centering
\footnotesize
\renewcommand{\arraystretch}{1.2}
\setlength{\tabcolsep}{8pt}
\begin{tabular}{@{}lccc@{}}
\toprule
\textbf{Skill} & \textbf{Success} & \textbf{Total} & \textbf{SR} \\
\midrule
\texttt{move\_to} & 27 & 35 & 77.1\% \\
\texttt{pick\_up\_from} & 28 & 29 & 96.5\% \\
\texttt{place\_in} & 11 & 13 & 84.6\% \\
\texttt{place\_on} & 6 & 6 & 100.0\% \\
\texttt{open\_door} & 7 & 7 & 100.0\% \\
\texttt{close\_door} & 4 & 6 & 66.7\% \\
\texttt{press} & -- & -- & -- \\
\texttt{turn\_on\_switch} & 7 & 8 & 87.5\% \\
\bottomrule
\end{tabular}
\caption{\textbf{Isolated VLA skill benchmark results from BEHAVIOR-1K skill instances.} Long-horizon deployment additionally requires robustness to chained initial conditions. Both protocols use the same skill-group checkpoints; the isolated benchmark serves them through the single-skill evaluation path, while the end-to-end harness routes them through the agent loop. Success is scored by the multi-view VLM verifier with human review of its decisions, rather than the simulator BDDL predicate, on validation-split snapshots drawn from the same instances evaluated end-to-end. Per-skill samples are small, so exact rates should not be over-interpreted; \texttt{press} shares the switch/press checkpoint with \texttt{turn\_on\_switch} and was not separately benchmarked in isolation.}
\label{tab:single-skill}
\end{table}

\subsection{Competence Does Not Survive Composition}

The single-skill and end-to-end protocols evaluate different semantic states. In the single-skill benchmark, a grasp skill starts with the object, robot pose, and camera view close to a successful demonstration boundary. In end-to-end execution, skill \(s_t\) starts from the actual terminal state produced by \(s_{t-1}\): shifted gripper poses, displaced objects, off-angle camera views, or ambiguous object instances. The result is a skill-to-task gap: VLA skills that are competent from snapshot states become unreliable when called from chained states. Crucially, both protocols route the \emph{same} skill-group checkpoints (Table~\ref{tab:single-skill}): although they use different success criteria---isolated skill success is VLM-judged with human review, composed task success is the BDDL predicate---the checkpoint that scores highly in isolation is exactly the policy that stalls in composition. The gap is not visible in isolated snapshot evaluation alone: because the same checkpoints are used in both protocols, the traces suggest that chained boundary states expose a robustness failure that is not explained by isolated skill success rates. We hypothesize that simply scaling the same snapshot-only training distribution would not fully close this gap; such scaling adds density on the curated boundary distribution, where the skills are already competent, rather than coverage of the chained states where the observed failures concentrate. Targeted chained-state data and readiness-aware verification are therefore likely necessary.

\subsection{Attributing Handoff Failures to Actionable Causes}
\label{sec:e2e-traces}

Table~\ref{tab:progress-score} reports a quantitative per-task signal over three instances of each task: a \emph{progress score} equal to the fraction of the task's reference skill sequence, the BEHAVIOR-1K teacher-hint plan, that a rollout completes, scored by hand from the rollout video. Because task-level predicate success is near zero, this graded score is what separates a run that stalls at its first navigation from one that clears most of its plan: mean progress across the thirty rollouts is 19.5\%, ranging from roughly half the reference plan on the radio and popcorn tasks down to zero on Easter eggs, and the score localizes where competence collapses---contact-rich manipulation and late multi-object steps.

\begin{table}[t]
\centering
\footnotesize
\renewcommand{\arraystretch}{1.2}
\setlength{\tabcolsep}{6pt}
\begin{tabular}{@{}lcc@{}}
\toprule
\textbf{Task} & \textbf{Ref.} & \textbf{Progress (\%)} \\
\midrule
Turn on radio          & 4  & 50.0 $\pm$ 43.3 \\
Make microwave popcorn & 8  & 45.8 $\pm$ 7.2 \\
Move boxes to storage  & 10 & 30.0 $\pm$ 0.0 \\
Pick up trash          & 12 & 22.2 $\pm$ 4.8 \\
Put shoes on rack      & 14 & 11.9 $\pm$ 4.1 \\
Bring in wood          & 13 & 10.3 $\pm$ 4.4 \\
Cook hot dogs          & 13 & 10.3 $\pm$ 4.4 \\
Set mousetraps         & 12 & 8.3 $\pm$ 0.0 \\
Freeze pies            & 26 & 6.4 $\pm$ 2.2 \\
Hide Easter eggs       & 12 & 0.0 $\pm$ 0.0 \\
\bottomrule
\end{tabular}
\caption{\textbf{Per-task progress score across the ten BEHAVIOR-1K tasks, three instances each.} The progress score is the fraction of the task's reference skill sequence (\textbf{Ref.}, the BEHAVIOR-1K teacher-hint plan) that a rollout completes. We score it by human inspection of the rollout video against the reference skill sequence: a step is counted complete only when the video shows its postcondition satisfied. Because alternative valid plans are not credited and scoring is a single annotator's judgement, this is a conservative diagnostic proxy rather than an official BEHAVIOR-1K metric. We report the mean $\pm$ standard deviation over the three instances ($n{=}3$). It recovers a graded signal where task-level predicate success is near zero: a rollout can clear most of its reference plan yet still miss the final task predicate (e.g.\ a completed radio sequence still leaves the appliance not registered as on). Progress concentrates in early navigation, door, and grasp steps and collapses at contact-rich manipulation and late multi-object placement.}
\label{tab:progress-score}
\end{table}

Beyond this aggregate progress score, the harness attributes every failed attempt to a specific cause. The three trace-derived causes---target grounding or scene search, control execution, and next-skill readiness (Table~\ref{tab:failure-taxonomy})---are all VLA-side; each points at a concrete skill and chained-state distribution to improve and maps to a targeted remedy. Crucially, all three are \emph{transfer} failures: the same checkpoints are competent in isolation (Table~\ref{tab:single-skill}), so the breakdown cannot be diagnosed from isolated skill success alone. This is the payoff of the agent layer: it turns near-zero task \emph{success} into a prioritized list of what the VLA must learn next. Notably, the binding failure at most handoffs is not premature advance through a weak state but the policy failing to \emph{reach} a ready state from its chained start, exhausting the skill budget. Tightening the navigation verifier to require an arm-reach rather than a merely same-region handoff is a controlled ablation that changes only the verifier criterion, holding tasks, instances, and checkpoints fixed. This sharpens the readiness signal: it surfaces 12 additional readiness failures and 25 more re-navigation attempts and recovers the radio task, while leaving control/commit the dominant cause (Table~\ref{tab:armreach-ablation}). A blinded human audit of the verifier (Appendix, Table~\ref{tab:verifier-audit}) confirms its failure decisions: of the verifier-flagged failures the annotators could adjudicate, 20 of 21 were judged real (over-strict $0.05$), so a reported failure is a real failure rather than a verifier artifact. Because the failure \emph{categories} remain verifier-derived and the audited set is small, we report the category counts as preliminary indicators rather than calibrated rates. Table~\ref{tab:failure-taxonomy} aggregates the failed-attempt categories; per-task attempt, verifier-call, and termination diagnostics appear in the appendix, drawn from a representative ten-rollout round, one instance per task, that issues 196 verifier calls in total.

\begin{table*}[t]
\centering
\footnotesize
\renewcommand{\arraystretch}{1.25}
\setlength{\tabcolsep}{5pt}
\begin{tabular}{@{}L{0.17\textwidth} L{0.13\textwidth} c L{0.35\textwidth} L{0.16\textwidth}@{}}
\toprule
\textbf{Attributed Cause} & \textbf{Sub-Category} & \textbf{n} & \textbf{Dominant Failure Signature} & \textbf{Targeted Remedy} \\
\midrule
\multirow{3}{=}{Control execution}
  & grasp & 31 & reaches the object but the grasp does not close or hold
  & \multirow{3}{=}{contact-rich, commit-frame data} \\
  & actuation & 15 & reaches the handle, switch, or button but it does not actuate & \\
  & placement & 12 & approaches the surface or container but mis-positions the object & \\
\midrule
Target grounding & scene search & 37 & target out of view or the wrong instance after navigation or door opening & instance-grounding supervision \\
\midrule
Next-skill readiness & navigation-to-ready & 35 & navigation stops short of an arm-reach pose at the handoff & chained-state fine-tuning \\
\bottomrule
\end{tabular}
\caption{\textbf{Failed skill attempts attributed to three VLA-side causes, with sub-category breakdown and remedy} (live VLA, $K{=}200$, a representative round, one official instance per task; 130 failed attempts). All three are \emph{transfer} failures of in-isolation-competent skills (Table~\ref{tab:single-skill}), not raw incapacity. Verifier-side cases (the two excluded attempts) are not counted here; counts derive from structured verifier reasons and are diagnostics, not ground-truth labels.}
\label{tab:failure-taxonomy}
\end{table*}

\section{Discussion}

\subsection{Implications}

The results support reading the agent layer as a diagnostic instrument for long-horizon robotics, and should be read as preliminary diagnostics. The harness executes, verifies, and records each rollout, turning a budget-limited run into a step-resolved trace of where execution stalls. These traces reveal concrete blocked states where the remaining issue is most often the chained-state distribution the VLA was never trained on---a skill-boundary distribution shift---rather than a uniformly weak policy; target grounding is the other recurring factor. Further ablations that rerun the same task suite while varying handoff checks, verifier cadence, or recovery strength would measure whether the bottleneck moves; we leave these to future work.

\subsection{Limitations}

The current end-to-end evidence covers a small number of completed pilot traces, so bottleneck counts should not be read as population-level estimates. The verifier is VLM-based; a blinded two-annotator audit (Table~\ref{tab:verifier-audit}) confirms its failure decisions (20 of 21 verifier-flagged failures judged real, over-strict $0.05$), but it validates only the binary failure decision---the failure-category labels remain verifier-derived and may carry bias for object placement, grasp visibility, and articulated-object state, and the audited set is small. The task scripts use some under-specified object names such as \texttt{door} or \texttt{storage box}; robust instance-level binding remains open. Finally, recovery is handled by the agent replanning within the episode budget---in practice it usually re-navigates to re-establish the handoff before retrying a failed skill. A no-recovery versus replanning ablation, runnable in our harness, is designed to separate VLA skill brittleness from recovery limitations; an oracle skill-boundary reset remains a future upper bound because our current live execution setup does not expose arbitrary skill-start state restoration.

\subsection{Future Work: Next-Skill Readiness Verification}
\label{sec:future}

The handoff failures diagnosed above motivate a principled treatment of next-skill readiness (Sec.~\ref{sec:handoff}). Rather than hand-authoring an arm-reach clause into each \texttt{move\_to} postcondition, the verifier could check
\[
\phi_t \wedge \rho(s_t, s_{t+1}),
\]
where \(\rho(s_t, s_{t+1})\) is a next-skill readiness predicate generated from the type and arguments of \(s_{t+1}\):
\[
\rho(s_t, s_{t+1}) = R_{\text{name}(s_{t+1})}(\text{args}(s_{t+1}), o_t),
\]
\(o_t\) is the current multi-view visual evidence and \(R\) is a finite template library indexed by the next skill type. For example, a \texttt{move\_to} before \texttt{pick\_up\_from} would pass only when the target is plausibly graspable, whereas one before \texttt{open\_door} would instead require the handle within reach (Table~\ref{tab:readiness-templates}). Evaluating this design---in particular a postcondition-only versus handoff-aware ablation that isolates the effect of \(\rho\)---requires a verifier conditioned on the next skill and is left to future work.

\section{Related Work}

\textbf{Robot policy learning.}
Language-conditioned manipulation has progressed from modular imitation-learning systems such as CLIPort and PerAct~\citep{shridhar2022cliport,shridhar2022peract} and prompt-conditioned agents such as VIMA~\citep{jiang2022vima} to large robot policies and VLA models including Gato, RT-1, RT-2, Open X-Embodiment, Octo, OpenVLA, \(\pi_0\), and diffusion policies~\citep{reed2022gato,brohan2022rt1,brohan2023rt2,oneill2023openx,team2024octo,kim2024openvla,black2024pi0,chi2023diffusion}. Recent work further strengthens the VLA substrate through world models and learned memory~\citep{cen2025worldvla,shi2025memoryvla}. We build a compact \(\pi_{0.5}\)-based VLA skill library from cleaned BEHAVIOR-1K skill data, but focus on the execution problem around that library: how trained skills should be composed, verified, recovered, and diagnosed under chained initial conditions.

\textbf{Task and motion planning and programmatic policies.}
Task and motion planning combines symbolic task structure with continuous motion, grasp, placement, and collision constraints~\citep{kaelbling2011hierarchical,garrett2020pddlstream}; code-generating approaches such as Code as Policies synthesize executable robot programs over perception and control APIs~\citep{liang2023code}. MAESTRO and TIGeR similarly enrich VLM agents with active perception, geometry, planning, control, or exact geometric-computation tools~\citep{shi2025maestro,han2025tiger}. H-WM guides task and motion planning with a hierarchical world model that curbs error accumulation across long-horizon VLA execution~\citep{huang2026hwm}. These systems show the value of stable tools with inspectable semantics. Our harness treats trained VLA skills as agent-callable tools and supplies a missing runtime interface: typed skill contracts, handoff-sensitive postcondition checks, and trace-backed failure attribution.

\textbf{Language-conditioned agents and semantic grounding.}
SayCan, Inner Monologue, and ReAct-style systems use language models to select, reason over, or call robot tools~\citep{ahn2022saycan,huang2022inner,yao2023react}. Recent agentic robot systems add structured workflows, multi-agent decomposition, temporal verification, unified data-collection/deployment loops, and stronger embodied VLM backbones~\citep{yang2025maniagent,yang2025agenticrobot,li2026roboclaw,robobrain2025}. Semantic grounding work also uses language or vision-language models to build spatial constraints, value maps, or reusable atomic skills~\citep{huang2023voxposer,li2025atomicskills}. We target execution-time grounding at skill handoffs: a step should advance not when the target is merely visible but when the state is ready for the next skill---a readiness notion we fold into the skill postcondition and check online rather than as an offline label.

\section{Conclusion}

We presented a semantic execution harness for agent-orchestrated VLA long-horizon skill composition. The harness represents skills through typed semantic contracts, executes VLA policies under bounded step-level verification, and surfaces semantic handoff failures at skill boundaries. Preliminary BEHAVIOR-1K diagnostics show a gap between snapshot skill competence and chained-state robustness. This gap is best interpreted as a roadmap for VLA improvement: the agent layer supplies task control, verification, recovery, and evidence, while trace-backed semantic diagnostics identify the VLA recovery states and grounding capabilities needed for reliable long-horizon robots.

\bibliographystyle{plainnat}
\bibliography{references}

\clearpage
\section*{Appendix}

\subsection*{Training and Routing Details}

The skills are \(\pi_{0.5}\) vision-language-action policies~\citep{black2024pi0}: a PaliGemma 2B vision-language backbone coupled with a 300M-parameter flow-matching action expert that consumes three \(224\times224\) RGB views from an egocentric head camera and two wrist cameras together with proprioceptive state and predicts multi-step action chunks. We clean BEHAVIOR-1K teleoperation demonstrations~\citep{li2023behavior} into skill segments aligned with the skill contracts using the dataset's fine-grained subtask annotations, with a per-task split of 180 training and 20 validation episodes. Training has two stages. A full-skill mid-training stage finetunes the base \(\pi_{0.5}\) checkpoint over the pooled skill data for 50k steps to preserve shared visuomotor and language grounding. From this initialization, skill-group post-training specializes separate checkpoints---navigation/orientation, switch/press, door open/close, grasping, surface placement, and containment placement---with full-parameter updates under a cosine learning-rate schedule (peak \(2.5\times10^{-5}\), global batch size 256). At deployment the agent routes each skill type to its group's checkpoint and renders it into the skill's language prompt (Table~\ref{tab:skill-checkpoints}).

\begin{table}[h]
\centering
\footnotesize
\renewcommand{\arraystretch}{1.2}
\setlength{\tabcolsep}{10pt}
\begin{tabular}{@{}lcc@{}}
\toprule
\textbf{Skill} & \textbf{Mean Frames} & \textbf{Mean Seconds} \\
\midrule
\texttt{move\_to} & 692 & 23.1 \\
\texttt{pick\_up\_from} & 598 & 19.9 \\
\texttt{place\_on} & 430 & 14.3 \\
\texttt{place\_in} & 346 & 11.5 \\
\texttt{press} & 291 & 9.7 \\
\texttt{open\_door} & 941 & 31.4 \\
\texttt{close\_door} & 505 & 16.8 \\
\bottomrule
\end{tabular}
\caption{\textbf{Skill-duration statistics from selected BEHAVIOR-1K demonstrations.} The spread motivates skill-specific budgets and bounded verification.}
\label{tab:skill-duration}
\end{table}

\begin{table}[h]
\centering
\footnotesize
\renewcommand{\arraystretch}{1.2}
\setlength{\tabcolsep}{6pt}
\begin{tabular}{@{}lL{0.32\columnwidth}cc@{}}
\toprule
\textbf{Skill} & \textbf{Checkpoint Group} & \textbf{Steps} & \textbf{Budget $B_s$} \\
\midrule
\texttt{move\_to} & nav./orient. & 60k & 800 \\
\texttt{press} & switch/press & 60k & 800 \\
\texttt{turn\_on\_switch} & switch/press & 60k & 800 \\
\texttt{open\_door} & door open/close & 60k & 1500 \\
\texttt{close\_door} & door open/close & 60k & 800 \\
\texttt{pick\_up\_from} & grasping & 180k & 800 \\
\texttt{place\_on} & surface placement & 180k & 800 \\
\texttt{place\_on\_next\_to} & surface placement & 180k & 1200 \\
\texttt{place\_in} & containment placement & 200k & 800 \\
\bottomrule
\end{tabular}
\caption{\textbf{Skill-to-checkpoint routing and per-skill budgets used in all experiments.} All checkpoints are $\pi_{0.5}$ models finetuned from a shared 50k-step all-skills initialization; skills in the same group share a checkpoint.}
\label{tab:skill-checkpoints}
\end{table}

\subsection*{Additional Diagnostics}

This appendix collects the per-rollout diagnostics behind the main-text attribution.

Table~\ref{tab:armreach-ablation} reports the effect of tightening the navigation handoff criterion to an arm-reach test, which reclassifies handoff stalls and recovers the radio task.

\begin{table}[tbp]
\centering
\footnotesize
\renewcommand{\arraystretch}{1.25}
\setlength{\tabcolsep}{6pt}
\begin{tabular}{@{}L{0.50\columnwidth} c c@{}}
\toprule
\textbf{Diagnostic} & \textbf{Reached-area} & \textbf{Arm-reach} \\
\midrule
Tasks solved ($q{=}1$)            & 0 / 10 & 1 / 10 \\
Mean task score $\bar q$          & 0.01   & 0.10 \\
\midrule
\texttt{move\_to} attempts    & 74     & 99 \\
Next-skill readiness failures     & 23     & 35 \\
Target grounding failures         & 31     & 37 \\
Control / commit failures         & 61     & 58 \\
Total failed attempts             & 115    & 130 \\
\bottomrule
\end{tabular}
\caption{\textbf{Effect of the arm-reach handoff criterion in the navigation verifier} (live VLA, $K{=}200$, a representative round, one official instance per task). Both columns use the same tasks, instances, and checkpoints; only the navigation verifier criterion differs (reached-area vs.\ arm-reach), making this a controlled ablation. Tightening the navigation postcondition from \emph{reached the target area} to \emph{the target is within arm reach} reclassifies handoff stalls: it surfaces 12 additional next-skill readiness failures and triggers 25 more \texttt{move\_to} attempts (re-navigation to a reachable pose), while control/commit execution stays the binding cause. It also recovers the radio task, whose toggle only succeeds once the robot is forced to re-navigate to a reachable pose. Counts are trace-derived diagnostics from structured verifier reasons (cf.\ Table~\ref{tab:failure-taxonomy}), not ground-truth labels.}
\label{tab:armreach-ablation}
\end{table}

The completed ablation is the arm-reach handoff criterion (Table~\ref{tab:armreach-ablation}). Three further ablations run on the same task suite, instances, and checkpoints but are left to future work: a no-recovery versus replanning comparison that separates VLA skill brittleness from recovery-policy limits, a verifier-cadence sweep over \(K \in \{50,100,200\}\), and a next-skill-conditioned readiness predicate that verifies \(\phi_t \wedge \rho\) directly rather than through \texttt{move\_to} wording (Sec.~\ref{sec:future}).

Table~\ref{tab:trace-diagnostics-10} gives the per-task attempt, success, verifier-call, and termination counts behind these diagnostics.

\textbf{Verifier example.} The verifier is queried with the skill's natural-language postcondition, plus an arm-reach clause for \texttt{move\_to}. On the radio rollout, the first \texttt{move\_to(radio)} check, logged under the raw token \texttt{navigate\_to}, asks: \emph{``Has the robot successfully completed this skill: `navigate to the radio'? Judge only from the current camera views. For navigate\_to, success requires the navigation target to appear close-up and prominent in the head camera at arm-reach distance; if it is only visible far away, small, or across the room, answer no.''} It returns \texttt{verdict=no}, \texttt{confidence=0.8}, with rationale \emph{``the head camera does not show any object identifiable as a radio close-up and prominent at arm-reach distance\ldots\ the robot has not successfully navigated to the radio,''} and \texttt{recovery\_hint} \emph{``re-localize or search for the radio.''} This readiness verdict triggers re-navigation, and the task later succeeds once the radio is reached at arm-reach---the re-plan/retry/advance loop visualized in Fig.~\ref{fig:teaser}.

\textbf{Verifier audit.} Because the failure attribution is read off the verifier, we audit its failure decisions. Two authors independently judged sampled verifier decisions from the same head and wrist views, blind to the verifier verdict, agreeing on 90\% of non-\emph{unclear} items. Of the verifier's failure decisions they could adjudicate in consensus (21), 20 were confirmed as real failures---an over-strict rate of $0.05$, so the verifier almost never rejects a skill that actually succeeded and a reported failure is a real failure. Task-level success is independent of this audit, being scored by the BDDL predicate. The audit covers the binary failure decision only, not the failure category, and $N$ is small (Table~\ref{tab:verifier-audit}).

\begin{table}[t]
\centering
\footnotesize
\renewcommand{\arraystretch}{1.2}
\begin{tabular}{@{}lc@{}}
\toprule
\textbf{Metric} & \textbf{Value} \\
\midrule
Audited verifier failures (2-annotator consensus) & 21 \\
Confirmed real failures (human) & 20 \\
Failure-decision reliability & 0.95 \\
Over-strict rate (real success wrongly failed) & 0.05 \\
Inter-annotator agreement & 90\% \\
\bottomrule
\end{tabular}
\caption{\textbf{Blinded human audit of the verifier's failure decisions.} Two authors independently judged sampled verifier decisions from the same head/wrist views, blind to the verdict; they agreed on 90\% of non-\emph{unclear} items. Of the verifier's \emph{failure} decisions they could adjudicate (21 in consensus)---the decisions that underlie the failure attribution in Table~\ref{tab:failure-taxonomy}---20 were confirmed as real failures, an over-strict rate of $0.05$. A reported failure is thus very likely a real failure rather than a verifier artifact. The audit validates the binary failure decision, not the failure \emph{category}, which remains verifier-derived; $N$ is small and these are preliminary indicators.}
\label{tab:verifier-audit}
\end{table}

\begin{table*}[p]
\centering
\footnotesize
\renewcommand{\arraystretch}{1.2}
\setlength{\tabcolsep}{6pt}
\begin{tabular}{@{}L{0.24\textwidth}ccccccc@{}}
\toprule
\textbf{Task} & \textbf{Attempts} & \textbf{Succ.} & \textbf{Fail} & \textbf{Frames} & \textbf{VLM Calls} & \textbf{End} & \textbf{$q$} \\
\midrule
Turn on radio & 5 & 3 & 2 & 2\,160 & 4 & success & 1 \\
Pick up trash & 20 & 9 & 11 & 10\,535 & 16 & timeout & 0 \\
Make microwave popcorn & 14 & 9 & 5 & 6\,475 & 12 & timeout & 0 \\
Move boxes to storage & 44 & 13 & 31 & 27\,900 & 39 & finish & 0 \\
Set mousetraps & 27 & 9 & 18 & 16\,000 & 23 & finish & 0 \\
Hide Easter eggs & 23 & 9 & 14 & 10\,800 & 21 & finish & 0 \\
Freeze pies & 33 & 15 & 18 & 23\,180 & 32 & finish & 0 \\
Cook hot dogs & 28 & 15 & 13 & 18\,289 & 22 & timeout & 0 \\
Bring in wood & 16 & 15 & 1 & 7\,100 & 10 & finish & 0 \\
Put shoes on rack & 22 & 3 & 19 & 15\,384 & 17 & timeout & 0 \\
\midrule
Total (10 tasks) & 232 & 100 & 132 & 137\,823 & 196 & --- & 1/10 \\
\bottomrule
\end{tabular}
\caption{\textbf{Per-task end-to-end trace diagnostics} (live VLA, $K=200$, a representative round, one official instance per task). All ten BEHAVIOR-1K tasks supported by the current skill library are covered with one rollout each. Succ.\ and Fail count skill attempts (not task outcomes); Frames is the total environment steps consumed across all attempts; VLM calls is the cumulative verifier query count; End is the rollout termination reason (\texttt{success}/\texttt{timeout}/\texttt{finish}) and $q$ the BEHAVIOR-1K task-predicate score. Radio is the one solved task under the arm-reach handoff criterion; of the 132 raw failed attempts, two are verifier-side cases excluded from the attribution in Table~\ref{tab:failure-taxonomy} (130).}
\label{tab:trace-diagnostics-10}
\end{table*}

\subsection*{Proposed Readiness Template Library}

Table~\ref{tab:readiness-templates} lists the proposed next-skill readiness predicate templates \(R_{\text{name}(s_{t+1})}\) for instantiating \(\rho(s_t,s_{t+1})\) (Sec.~\ref{sec:future}). Each is indexed by the next skill type and is a skill-contract rule, not a per-task pairwise hand-written condition; the current system approximates them through \texttt{move\_to} postcondition wording rather than verifying \(\phi_t \wedge \rho\) directly.

\begin{table*}[p]
\centering
\footnotesize
\renewcommand{\arraystretch}{1.3}
\setlength{\tabcolsep}{6pt}
\begin{tabular}{@{}L{0.26\textwidth}L{0.64\textwidth}@{}}
\toprule
\textbf{Next Skill} & \textbf{Readiness Predicate Template} \\
\midrule
\texttt{pick\_up\_from(x,y)} & \(\mathrm{visible}(x) \wedge \mathrm{graspReachable}(x)\); merely visible at a distance is insufficient. \\
\texttt{open\_door(a)} & \(\mathrm{visible}(\mathrm{handleOrEdge}(a)) \wedge \mathrm{manipReachable}(a)\). \\
\texttt{close\_door(a)} & \(\mathrm{doorOpen}(a) \wedge \mathrm{visible}(\mathrm{handleOrEdge}(a)) \wedge \mathrm{manipReachable}(a)\). \\
\texttt{press(a)} & \(\mathrm{visible}(\mathrm{controlRegion}(a)) \wedge \mathrm{manipReachable}(a)\). \\
\texttt{turn\_on\_switch(a)} & \(\mathrm{visible}(\mathrm{controlRegion}(a)) \wedge \mathrm{manipReachable}(a)\). \\
\texttt{place\_in(x,c)} & \(\mathrm{held}(x) \wedge \mathrm{containerReachable}(c) \wedge \mathrm{openingPlausible}(c)\). \\
\texttt{place\_on(x,s)} & \(\mathrm{held}(x) \wedge \mathrm{supportReachable}(s)\). \\
\texttt{place\_on\_next\_to(x,s)} & \(\mathrm{held}(x) \wedge \mathrm{supportReachable}(s)\). \\
\texttt{move\_to(x)} & no additional predicate beyond \(\phi_t\); subsequent manipulation steps impose their own readiness predicate. \\
\bottomrule
\end{tabular}
\caption{\textbf{\emph{Proposed} next-skill readiness templates for instantiating \(\rho(s_t,s_{t+1})\) (Sec.~\ref{sec:future}).} They are skill-contract templates, not per-task pairwise hand-written rules. In the current system readiness is approximated through \texttt{move\_to} postcondition wording rather than these templates; verifying \(\phi_t\wedge\rho\) with a next-skill-conditioned verifier is future work.}
\label{tab:readiness-templates}
\end{table*}

\end{document}